%% file: paper.tex
\documentclass{llncs}
\usepackage{epsf}
\usepackage{array}
\usepackage{float}
\usepackage{latexsym}
\usepackage{amsmath}
\usepackage{graphicx}
\usepackage{mathtools}
\usepackage{burcu}
\newcommand{\argmax}{\arg\!\max}


\begin{document}

\title{A Trie-Structured Bayesian Model for Unsupervised Morphological Segmentation}
\author{Murathan Kurfal{\i}\inst{1} \and Ahmet \"{U}st\"{u}n\inst{1} \and Burcu Can\inst{2}}
\authorrunning{Kurfal{\i} et al.} % abbreviated author list (for running head)
%
%%%% list of authors for the TOC (use if author list has to be modified)
\tocauthor{Murathan Kurfal{\i}, Ahmet \"{U}st\"{u}n, Burcu Can}
\institute{Cognitive Science Department, Informatics Institute \\ Middle East Technical University (ODT\"U) \\
Ankara, 06800, Turkey\\
\email{\{kurfali,ustun.ahmet\}@metu.edu.tr}
\and
Department of Computer Engineering, Hacettepe University \\ 
Beytepe, Ankara, 06800, Turkey \\
\email{burcucan@cs.hacettepe.edu.tr}
}

\maketitle

\begin{abstract}
In this paper, we introduce a trie-structured Bayesian model for unsupervised morphological segmentation. We adopt prior information from different sources in the model. We use neural word embeddings to discover words that are morphologically derived from each other and thereby that are semantically similar. We use letter successor variety counts obtained from tries that are built by neural word embeddings. Our results show that using different information sources such as neural word embeddings and letter successor variety as prior information improves morphological segmentation in a Bayesian model. Our model outperforms other unsupervised morphological segmentation models on Turkish and gives promising results on English and German for scarce resources.   
\end{abstract}

\section{Introduction}
Morphological segmentation is the task of segmenting words into their meaningful units called \textit{morphemes}. For example, the word \textit{transformations} is split into \textit{trans}, \textit{form}, \textit{ation}, and \textit{s}. This process serves mainly as a preprocessing task in many natural language processing (NLP) applications such as information retrieval, machine translation, question answering, etc. This process is essential because sparsity becomes crucial in those NLP applications due to morphological generation that produces various word forms from a single root. It is infeasible to build a dictionary that involves all possible word forms in a language in order to use in an NLP application.  Hankamer~\cite{Hankamer} suggests that the number of possible word forms in an agglutinative language such as Turkish is infinite. Therefore, instead of building a model based on word forms, morphological segmentation is applied to reduce the sparsity principally in any NLP application. 

Various features have been used for morphological segmentation. Many approaches use orthographic features. However, morphology is tightly connected with syntax and semantics. Syntactic and semantic features have also been used for the segmentation task.

Features are normally used in Bayesian models in the form of a prior distribution. For example, \cite{CreutzLengthFrequency} utilize frequency and length information of morphemes as prior information, which provide some orthographic features. 

In this paper, we aggregate prior information from different sources in morphological segmentation within a Bayesian framework. We use orthographic features such as letter successor variety (LSV) counts obtained from tries, semantic information obtained from the neural word embeddings~\cite{Mikolov} to measure the semantic relatedness between substrings of a word, and we use the presence information of a stem in a dataset after its suffixes are stripped off assuming a concatenative morphology. Our results show that combining prior information from different sources give promising results in unsupervised morphological segmentation. 

In this study, we learn tries based on semantic and orthographic features. Therefore, the output of our model is not only segmentation, but also tries that are composed of semantically and morphologically related words. 

The paper is organized as follows: Section~\ref{previouswork} presents the previous work on unsupervised morphological segmentation, section~\ref{modeldefinition} defines the mathematical model, section~\ref{inference} describes the inference algorithm to learn the mathematical model, section~\ref{experiments} presents the experimental results, and finally section~\ref{conclusion} concludes the paper with a discussion and potential future work. 

\section{Related Work}
\label{previouswork}

\begin{figure*}[t!]
  \centering
  \includegraphics[width=0.7\textwidth]{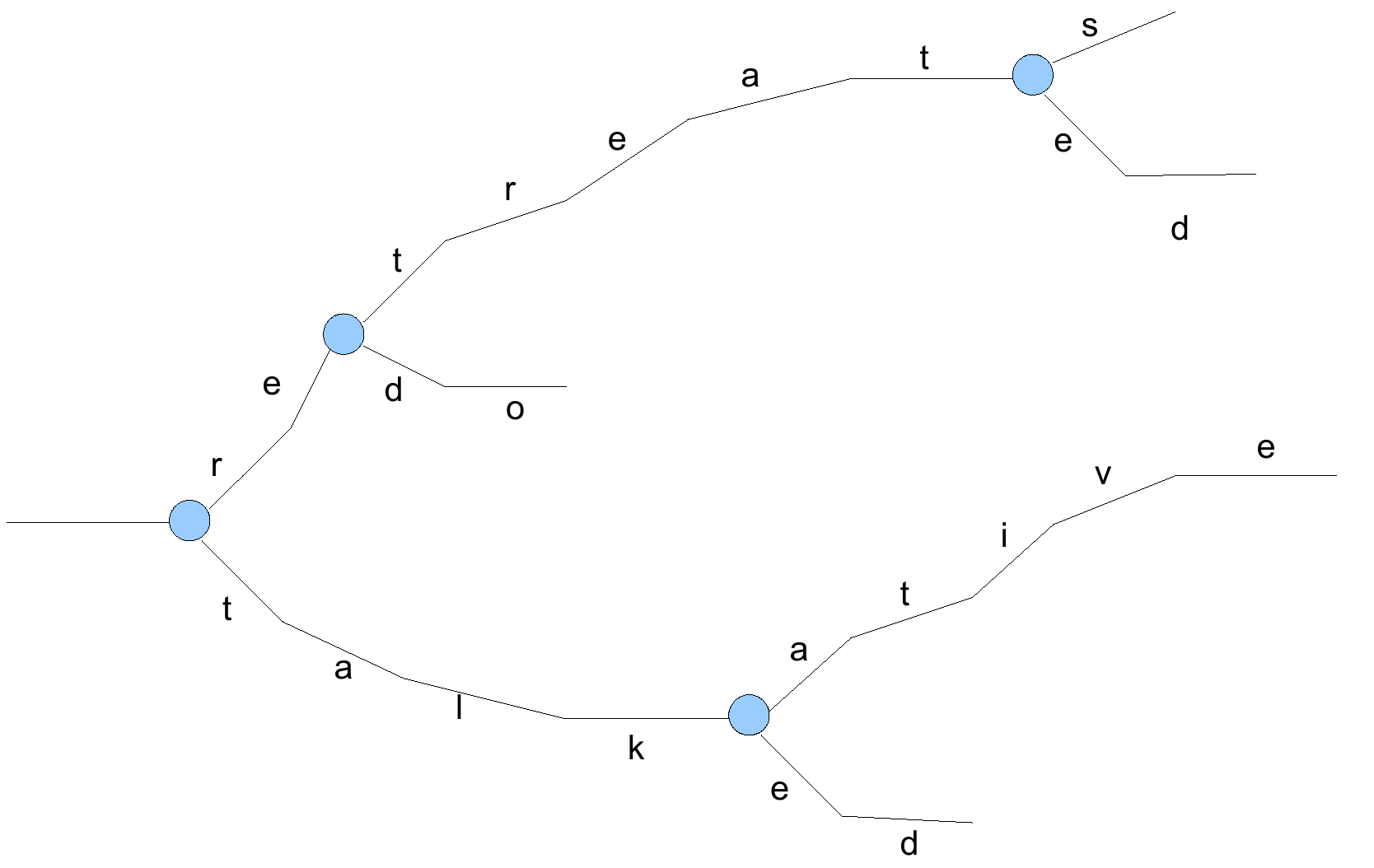}
  \caption{Potential morpheme boundaries on a trie used in LSV model~\cite{BurcuThesis}}
  \label{fig:lsv}
\end{figure*}

Morphological segmentation, as one of the oldest fields in NLP, has been excessively studied. Deterministic methods are the oldest ones used in morphological segmentation. Harris~\cite{1955Harris} defines the distributional characteristics of letters in a word for the first time for unsupervised morphological segmentation. LSV model is named after Harris, which defines the morpheme boundaries based on letter successor counts. If words are inserted into a trie, branches correspond to potential morpheme boundaries. An example is given in Figure~\ref{fig:lsv}. In the example, \textit{re-} is a potential prefix, and \textit{-s}, \textit{-ed} and \textit{-ing} are 
potential suffixes in the trie due to branching that emerges before those morphemes.  
LSV model has been applied in various works \cite{HaferWeiss,Dejean,Bordag2005,Bordag2006,Bordag2008}. In our study, we also use a LSV-inspired prior information, but this time in a Bayesian framework.

Stochastic methods have also been extensively used in unsupervised morphological segmentation. Morfessor is the name of the family of a group of unsupervised morphological segmentation systems which are all stochastic \cite{CreutzBaseline,Creutz2007,CreutzCategoriesMAP}. Non-parametric Bayesian models have also been applied in morphological segmentation \cite{GoldwaterInterpolatingBetween,Snyder2008,CanManandharEACL}. 

Neural-inspired features are used in the recent studies. Narasimhan et al.~\cite{Karthik} use semantic similarity obtained from neural word embeddings by word2vec~\cite{Mikolov}. Narasimhan et al.~\cite{Karthik} adopt the semantic similarity as a feature in a log-linear model. Soricut and Och~\cite{Soricut} use word embeddings to learn morphological rules in an unsupervised setting. 

In this work, we are both inspired by the oldest works and the recent works in terms of various features used. Thus our model has inspirations from the LSV methods, stochastic methods, and neural-based models in a combined framework. 

\section{Building Neural Word Embedding-based Tries}
\label{modeldefinition}

Our model is based on neural word embedding-based tries that are built by using two different methods:

\subsection{Tries Structured from the Same Stem}\label{recTrie}

These tries contain semantically related (morphologically derived or inflected from each other) words having the same stem. In order to find the stem of a given word in the training set, we used the algorithm which is introduced in \cite{ustun2016unsupervised}. In the algorithm, all potential prefixes of a word are extracted. For example, \textit{fe}, \textit{fea}, \textit{fear}, \textit{fearf}, \textit{fearfu}, \textit{fearful},  \textit{fearfull}, \textit{fearfully} are the prefixes of \textit{fearfully}. The rightmost segmentation point where the cosine similarity between the word and the first prefix (from the right of the word; i.e. \textit{fearful}) is higher than a manually set threshold\footnote{We assign 0.25 as the threshold following \cite{ustun2016unsupervised}.} gives the first valid prefix, which refers to the first segmentation point. 

Other segmentation points are found by repeating the process towards the head of the word by checking the cosine similarity between the just detected valid prefix and the subsequent prefix to the left of the word. The final detected prefix with the leftmost segmentation point in the word becomes the stem of the word. 

%After each possible prefix is pushed to the stack, a prefix is popped from the stack and cosine similarity between it and the control item, which is the given word in the first iteration is calculated. If cosine similarity is smaller than the threshold value\footnote{We assign 0.25 as the threshold following \cite{ustun2016unsupervised}.}, comparison continues with the subsequent prefix in the stack. Otherwise, the control item is replaced with the compared prefix. This process continues until the stack becomes empty and the last control item is marked as the stem \footnote{Since we are using a stack, the first prefix we consider is also the longest prefix while the second one is the second longest and so on. Therefore, the last control item is the shortest possible prefix with respect to algorithm.}. 

%% All prefixes of a given word are extracted in order to find the potential stem of the word. For example, \textit{fe}, \textit{fea}, \textit{fear}, \textit{fearf}, \textit{fearfu}, \textit{fearful}, \textit{fearfull}, \textit{fearfully} are all prefixes of \textit{fearfully}. Cosine similarity of each prefix and the current word is calculated. The shortest prefix having a cosine similarity over a threshold value\footnote{We assign 0.25 for the cosine similarity threshold in our study following \cite{ustun2016unsupervised}.} is marked as the stem. 

\begin{figure*}[t!]
  \centering
  \includegraphics[width=0.6\textwidth]{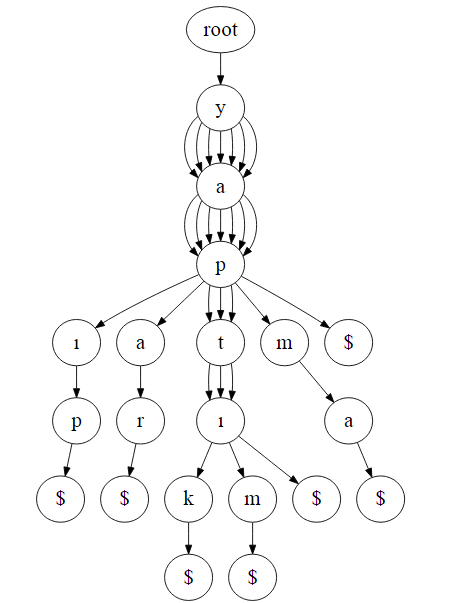}
  \caption{ Visualization of a trie portion that includes the word forms derived from the same stem. \textit{yap{\i}p}, \textit{yapar}, \textit{yapt{\i}k}, \textit{yapt{\i}m}, \textit{yapma} are inflected forms of the stem \textit{yap} (means \textit{to do}). The number of edges refers to the number of words in the corpus flowing in that direction on the trie. \$ denotes the end of the word.}
  \label{fig:trie_stem}
\end{figure*}

Among the nearest $50$ neighbors of the stem which are obtained from word2vec \cite{Mikolov}, the ones that begin with the same stem are inserted to the same trie. This process is repeated for each word that is inserted on the trie recursively until all the words that are semantically similar which share the same stem (detected by using the same algorithm described above) are covered. An example trie that is built with the words having the same stem is given in Figure \ref{fig:trie_stem}.

\subsection{Tries Based on Semantic Relatedness}\label{first50}

\begin{figure*}[t!]
  \centering
  \includegraphics[width=0.6\textwidth]{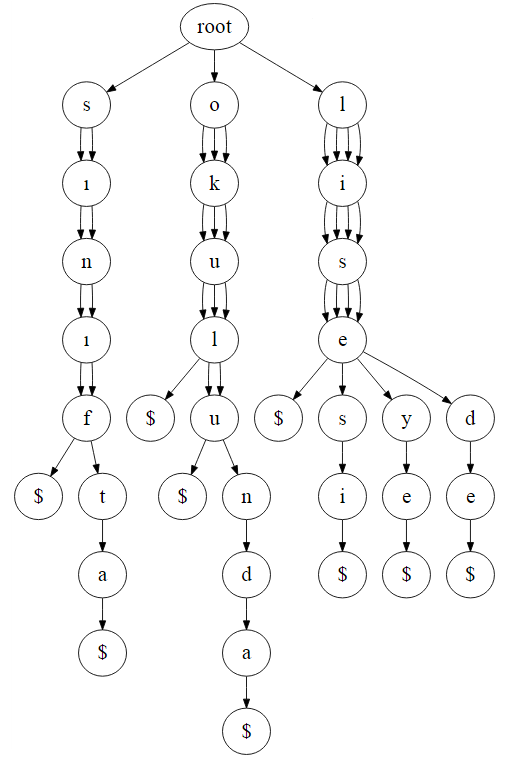}
  \caption{Visualization of a trie portion built by using semantic relatedness. The trie consists of the stems \textit{s{\i}n{\i}f}, \textit{okul}, \textit{lise} (means \textit{high school}, \textit{class}, \textit{school}) and affixed forms of these stems. The number of edges refers to the number of words in the corpus flowing in that direction on the trie. \$ denotes the end of the word.}
  \label{fig:trie_sr}
\end{figure*}

Semantically related 50 words are retrieved for each word in the training set by using word2vec~\cite{Mikolov}. For each word, a trie is built and 50 similar words are inserted on the word's trie. Eventually, a trie that consists of 51 words is created for each word in the training set. A portion of a trie that involves semantically related words is given in Figure \ref{fig:trie_sr}. 

\section{Bayesian Model Definition}

We define a Bayesian model in order to find the morpheme boundaries on the tries:
\begin{equation}
p(Model|Corpus) \propto p(Corpus|Model)p(Model)
\end{equation}
where $Corpus$ is a list of raw words and $Model$ denotes the segmentation of the corpus. The $Model$ that maximizes the given posterior probability will be searched for the segmentation task. We apply a unigram model for the likelihood:
\begin{eqnarray}
\label{trigram}
p(Corpus|Model) &=& \prod_i^{|W|}p(w_i=(m_{i1}+m_{i2}+\cdots+m_{it_i}|Model) \nonumber \\
                &=& \prod_i^{|W|} \prod_{j=1}^{t_i} p(m_{ij}|Model)
\end{eqnarray}
where $w_i$ is the ith word in $Corpus=\{w_1,\cdots,w_{|W|}\}$, $m_{ij}$ is the jth morpheme in $w_i$, $t_i$ is the number of morphemes in word $w_i$, and $|W|$ is the number of words in the corpus. Here, morphemes are generated by a Dirichlet Process (DP) as follows:
\begin{eqnarray}
m_{ij} &\propto& DP(\alpha,H) 
\end{eqnarray}
with the concentration parameter $\alpha$ and the base distribution $H$ that is formed with a geometric distribution:
\begin{eqnarray}
H(m_{ij}) &=& \gamma^{|m_{ij}|+1} 
\end{eqnarray}
Here, $|m_{ij}|$ is the length of $m_{ij}$ and $\gamma$ is the parameter of the geometric distribution. We assume that each letter is uniformly distributed. Therefore, we assign $\gamma=1/L$ where $L$ denotes the size of the alphabet in the language. Shorter morphemes will be favored with the usage of length-inspired base distribution in the DP. From the Chinese Restaurant Process (CRP) perspective, each morpheme is generated proportionally to the number of morphemes of the same type that have already been generated (i.e. customers having the same dish):
\begin{eqnarray}
p(m_{ij}=k|Model) = \frac{n_k+\alpha H(k)}{N+\alpha}
\end{eqnarray}
This computes the probability of $m_{ij}$ being of type $k$ where $k$ refers to  a distinct morpheme (i.e. morpheme type). Here, $n_k$ is the number of morphemes of type $k$ and $N$ is the total number of morpheme tokens in the model. We generate each morpheme regardless of its type, such as stem, prefix, or suffix. 

As for the prior information, we model the morpheme boundaries:
\begin{eqnarray}
p(Model) &=& \prod_i^{|W|} \prod_{j=1}^{t_i} p(b_{ij})
\end{eqnarray}
Here, $b_{ij}$ refers to the jth morpheme boundary in $w_i=m_{i1}+m_{i2}+\cdots+m_{it_i}$ where $w_i=\{b_{i1}, b_{i2}, \cdots, b_{it_i}\}$.

The probability of each $b_{ij}$ is decomposed in terms of the number of branches leaving that node (when inserted on the trie), semantic similarity that is introduced between the two word forms that is split with $b_{ij}$, and the presence of the word form once the suffix is stripped off from the word:
\begin{eqnarray}
\nonumber
p(b_{ij}) &=& p({b_{ij}}_{branch}) . p({b_{ij}}_{semantics}) . p({b_{ij}}_{presence})   
\end{eqnarray}
where $p({b_{ij}}_{branch})$ denotes the probability of $b_{ij}$ being a morpheme boundary based on the branches leaving the trie node, $p({b_{ij}}_{semantics})$ is based on the semantic similarity of the two word forms where $b_{ij}$ separates the two forms, and $p({b_{ij}}_{presence})$ is estimated based on the word form whether it exists in the corpus once the suffix after $b_{ij}$ is stripped off. 

Based on the LSV, the branching on the tries corresponds to the potential morpheme boundaries. We model the branching with a Poisson distribution:
\begin{eqnarray}
p({b_{ij}}_{branch}) &=&  p(z_{ij}=l|\lambda) \\
					 &=& \frac{\lambda^l e^{-\lambda}}{l!}
\end{eqnarray}
where $z_{ij}=l$ denotes the number of branches leaving the node below $b_{ij}$ and $\lambda$ is the parameter of the Poisson distribution\footnote{In the experiments, we assign $\lambda=4$.}. 

We use the cosine similarity (which is always between 0 and 1) between the neural word embeddings of the two word forms that are separated by $b_{ij}$ for the semantic distribution:
\begin{eqnarray}
p({b_{ij}}_{semantics}) &=& cos(x_{m_{i1}+\cdots+m_{ij}}, x_{m_{i1}+\cdots+m_{ij+1}})
\end{eqnarray}
Here, $x_{m_{i1}+\cdots+m_{ij}}$ corresponds to the word vector of the word form $m_{i1}+\cdots+m_{ij}$ obtained from word2vec. It is the full word vector and not the compositional vector obtained from morpheme vectors. 

As for the presence of the word form in the word list, we compute the likelihood of the word form $m_{i1}+\cdots+m_{ij}$:
\begin{eqnarray}
p({b_{ij}}_{presence}) &=& \frac{f(m_{i1}+\cdots+m_{ij})}{\sum_{i=1}^{|Corpus|}{f(w_i)}}
\end{eqnarray}
where $f(m_{i1}+\cdots+m_{ij})$ denotes the frequency of the word form in the corpus. 

\section{Inference}
\label{inference}

We use  Gibbs sampling~\cite{casella1992} for the inference. In each iteration, a word is uniformly selected from any trie and removed from the corpus. A binary segmentation of the word is sampled from the given posterior distribution:
\begin{align}
\MoveEqLeft[8] p(w_i=m_{i1}+m_{i2}|Corpus^{-w_i},Model^{-w_i},\alpha,\lambda,\gamma) \nonumber \\ 
&\propto p(m_{i1}|Model^{-w_i},\alpha,\gamma) p(m_{i2}|Model^{-w_i},\alpha,\gamma) p(b_{i1})  \\ \nonumber
\end{align}
Once a binary segmentation is sampled, another binary segmentation is sampled for $m_{i1}$. Therefore, a left-recursion is applied for the left part of the word. This is because of the cosine similarity that is computed between neural word embeddings of word forms and not suffixes by the original word2vec. 

\begin{figure*}[t!]
  \centering
  \includegraphics[width=0.75\textwidth]{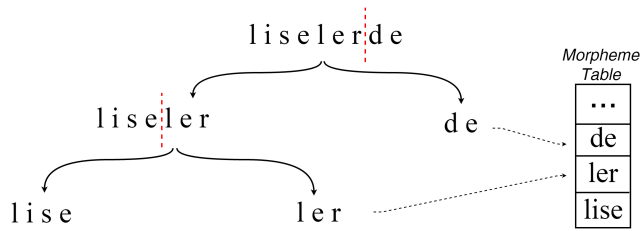}
  \caption{The binary segmentation of the word \textit{liselerde} (means \textit{in the high schools})}
  \label{fig:gibbs}
\end{figure*}

This process is repeated recursively until having at least 4 letters in the stem or having sampled the word itself from the posterior distribution (i.e. when the word is not segmented). An illustration is given in Figure \ref{fig:gibbs}. 

\section{Segmentation} \label{segmentation}

Once the model is learned, any unseen word can be segmented by using the learned model. Each word is split based on the maximum likelihood in the learned model:
\begin{equation}
\argmax_{m_{i1},\cdots,m_{it_i}} p(w_i=m_{i1}+\cdots+m_{it_i}|Model,\alpha,\gamma)  \\ 
\end{equation}

For the segmentation, we apply two different strategies. In both methods, we select the segmentation with the maximum likelihood, however the set of possible segmentations for the given word differs. In the first method, we only consider the segmentations learned by the model. Since the same word can exist in multiple tries, a word may have more than one different segmentation. In the second method, we consider all possible segmentations of a word and choose the one with the maximum likelihood. 

\section{Experiments and Results}
\label{experiments}

We did experiments on Turkish, English and German. For each language, we built two sets of tries based on the methods described in Section~\ref{recTrie} and Section~\ref{first50}. We aggregated the publicly available training and development sets provided by Morpho Challenge 2010~\cite{MorphoChallengeWeb} for English, Turkish and German for training. Although gold segmentations are provided in the datasets, we only used the raw words in training. Gold segmentations were only used for evaluation purposes. 

We began with 1686 English words, 1760 Turkish words, and 1779 German words obtained from the aggregated sets. Once the tries have been built by recursively augmenting the tries by using word2vec\cite{Mikolov}, eventually we obtained 2560 English word types, 43884 Turkish word types, and 13747 German word types in the tries structured from similar stems (see Section~\ref{recTrie}). Additionally, we obtained 34594 English word types, 67292 Turkish word types, and 23875 German types in the tries that were built based on the semantic relatedness (see Section~\ref{first50}). 

\begin{table}[t!]
\caption{Size of the datasets used in the experiments. \textit{m1} denotes the train set built by the first method (Section~\ref{recTrie}) and \textit{m2} denotes the train set built by the second method (Section~\ref{first50})}
\begin{center}
    \begin{tabular}{  l  l  l l  l } \hline
    \textbf{Language} & \textbf{Train-m1} & \textbf{Train-m2}   &\textbf{Train word2vec} & \textbf{Test} \\ \hline
    Turkish  & 43884 types & 67292 types & 725K types &  1760 types\\ 
    		 &  &  & 361M tokens &  \\ 
    English & 2560 types & 34594 types & 218K types & 1686 types\\ 
    		&   &   & 129M tokens  &  \\ 
    German  & 13747 types &  23875 types & 608K types  & 1779 types \\
    		& & & 651M tokens  & \\ \hline 
    \end{tabular}
\end{center}
\label{datatable}
\end{table}

We used 200-dimensional word embeddings that were obtained by training word2vec \cite{Mikolov} on 361 million word tokens and 725.000 word types in Turkish, 129 million word tokens and 218.000 word types in English, and 651 million word tokens and 608.000 word types in German. The size of all datasets used in the experiments are given in Table~\ref{datatable}.

We compared our model with Morfessor Baseline~\cite{CreutzBaseline} (M-Baseline), Morfessor CatMap~\cite{CreutzCategoriesMAP} (M-CatMAP) and MorphoChain System~\cite{Karthik}. For that purpose, we trained these models on the same training sets. We obtained the frequency information from the full word lists provided by Morpho Challenge which was need by other systems. The evaluation was performed on the aggregated training and development sets of Morpho Challenge 2010 using the Morpho Challenge evaluation method~\cite{MorphoChallengeWeb}. All word pairs that have a common morpheme are extracted from the results and checked whether they really share a common morpheme in the gold standard data. One point is given for each correct pair. The Precision is the proportion of the collected points to the total number of words. Recall is computed analogously. This time all word pairs that share a common morpheme are extracted from the gold standard data and checked whether they have a common morpheme in the results. For each correct pair, one point is given. Finally, the Recall is the proportion of the collected points is to the total number of words.

\begin{table}[!t]
\caption{Results obtained from the tries based on semantic relatedness (see  Section~\ref{first50}). TST denotes our trie-structured model.}
\centering
\begin{tabular}{lccc}

                  & \multicolumn{3}{c}{TURKISH}                         \\ \cline{2-4} 
                  & Precision (\%) & Recall (\%) &F-measure (\%) \\
TST &     58.27 &  \textbf{35.55} & \textbf{44.16}\\
M-CatMAP &  77.78  & 22.91 & 35.40  \\ 
M-Baseline & \textbf{84.39}  & 19.27 & 31.38  \\ 
MorphoChain  &  69.45   & 18.29 & 28.95 \\ 

                  & \multicolumn{3}{c}{ENGLISH}                         \\ \cline{2-4} 
M-Baseline &  64.82 & \textbf{64.07} & \textbf{64.44}   \\                  
TST &    56.40& 47.90 & 51.81\\ 
MorphoChain &   \textbf{86.26}  & 25.95 & 39.90 \\
M-CatMAP &  76.37   & 19.23  & 30.72  \\

                  & \multicolumn{3}{c}{GERMAN}                          \\ \cline{2-4}
M-Baseline  &  \textbf{64.74}   & 30.10  & \textbf{41.09}  \\ 
TST  &    38.66 &  \textbf{38.57} & 38.61      \\
M-CatMAP  &   62.32  &15.68 &25.06  \\ 
MorphoChain  &   56.39  &13.72 & 22.07 \\ 
\end{tabular}
\label{table:results}
\end{table}

\begin{table}[t!]
\caption{Results obtained from the tries structured from the same stem (see Section~\ref{recTrie}). TST denotes our trie-structured model.}
\centering
\begin{tabular}{lccc}
                  & \multicolumn{3}{c}{TURKISH}                         \\ \cline{2-4} 
                  & Precision (\%) & Recall (\%) &F-measure (\%) \\
M-CatMAP & 59.44  & \textbf{33.41} & \textbf{42.78} \\ 
TST     & 58.85  & 30.17& 39.89 \\ 
M-Baseline  & \textbf{74.09}  & 20.52 & 32.14 \\ 
Morpho-Chain  & 72.28  & 25.77 & 38.00  \\ 
                  & \multicolumn{3}{c}{ENGLISH}                         \\ \cline{2-4} 
                 
M-Baseline & \textbf{75.28}  & \textbf{61.05} &\textbf{67.42}  \\ 
TST  & 58.69  & 51.22& 54.70 \\ 
MorphoChain   & 91.74  & 30.39 & 45.66  \\ 
M-CatMAP    &  90.20  & 5.86  & 11.00  \\

                  & \multicolumn{3}{c}{GERMAN}                          \\ \cline{2-4}
M-Baseline  &59.65  & 29.47  & \textbf{39.45}    \\ 
TST   &  39.62 &  \textbf{35.28} &  37.33 \\ 
MorphoChain  & \textbf{79.06}  & 16.36 &  27.11 \\ 
M-CatMAP  & 55.96   & 16.41  & 25.38  \\
\end{tabular}
\label{table:results2}
\end{table}

The results are given in Table~\ref{table:results} and Table~\ref{table:results2} for tries that are composed of words structured from the same stem (see Section~\ref{recTrie}) and for tries that are based on semantic relatedness (see Section~\ref{first50}). According to the results, tries that contain semantically similar words achieve a better performance on morphological segmentation proving that semantically similar words also manifest similar syntactic and thus similar morphological features. 

Our trie-structured model (TST) performs better than Morfessor Baseline~\cite{CreutzBaseline}, Morfessor CatMAP \cite{CreutzCategoriesMAP} and Morphological Chain~\cite{Karthik} on Turkish with a F-measure of \%44.16 on the tries based on semantic relatednesss. We obtained a F-measure of \%39.89 for Turkish from the tries structured from the same stem, which is poorer than the other method. This shows that for morphologically rich languages, semantic relatedness plays a more important role in segmentation. That is because of the sparseness of the word forms in morphologically rich languages. Here we overcome the sparsity problem with semantic information that is used in semantically built tries. 

Our TST model performs better on the tries structured from the same stem on English with a F-measure of \%54.70 compared to the tries based on semantic relatedness, which has a F-measure of \%51.81. Since English is not a morphologically rich language, obtaining the correct stem plays an important role in segmenting the word. Words usually do not have more than one suffix and therefore finding the stem is normally sufficient to do morphological segmentation in morphologically poor languages such as English. 

Our German results are close to each other obtained from two types of tries. We obtain a F-measure of \%38.61 from the tries based on semantic relatedness and it performs better than Morfessor CatMAP and Morphological Chain. The F-measure is \%37.33 on German, which is obtained from the tries structured from the same stem. 

The results also show that Morfessor CatMAP suffers from sparsity in small datasets (especially in English), whereas our trie-structured model learns also well in small datasets.

\section{Conclusion and Future Work}
We propose a Bayesian model that utilizes semantically built trie structures that are built by using neural word embeddings (i.e. obtained from word2vec \cite{Mikolov}) for morphological segmentation in an unsupervised setting. The current study constitutes the first part of the on-going project which in the end aims to learn part-of-speech tags and morphological segmentation jointly. To this end, the fact that the tries having semantically related words achieves the best performance paves the way of using semantically similar words in learning syntactic features.

Moreover, considering the resource-scarce languages like Turkish, our trie-structured model shows a good performance on comparably smaller datasets. In comparison to other available systems, our model outperforms them in spite of the limited training data. This shows that the small size of data can be compensated to a certain extent with structured data, that is the main contribution of this paper. 
\label{conclusion}

\section*{Acknowledgments}
This research is supported by the Scientific and Technological Research Council of Turkey (TUBITAK) with the project number EEEAG-115E464. 

\bibliographystyle{splncs}
\bibliography{paper}
\end{document}

%% file: paper.bbl
\begin{thebibliography}{10}
\providecommand{\url}[1]{\texttt{#1}}
\providecommand{\urlprefix}{URL }

\bibitem{Bordag2005}
Bordag, S.: Unsupervised knowledge-free morpheme boundary detection. In:
  Proceedings of the RANLP 2005 (2005)

\bibitem{Bordag2006}
Bordag, S.: Two-step approach to unsupervised morpheme segmentation. In:
  Proceedings of 2nd Pascal Challenges Workshop. pp. 25--29 (2006)

\bibitem{Bordag2008}
Bordag, S.: Unsupervised and knowledge-free morpheme segmentation and analysis.
  Advances in Multilingual and Multimodal Information Retrieval pp. 881--891
  (2008)

\bibitem{BurcuThesis}
Can, B.: Statistical Models for Unsupervised Learning of Morphology and POS
  tagging. Ph.D. thesis, Department of Computer Science, The University of York
  (2011)

\bibitem{CanManandharEACL}
Can, B., Manandhar, S.: Probabilistic hierarchical clustering of morphological
  paradigms. In: Proceedings of the 13th Conference of the European Chapter of
  the Association for Computational Linguistics. pp. 654--663. EACL '12,
  Association for Computational Linguistics (2012)

\bibitem{casella1992}
Casella, G., George, E.I.: Explaining the {G}ibbs sampler. The American
  Statistician  46(3),  167--174 (1992)

\bibitem{CreutzLengthFrequency}
Creutz, M.: Unsupervised segmentation of words using prior distributions of
  morph length and frequency. In: Proceedings of the 41st Annual Meeting on
  Association for Computational Linguistics. pp. 280--287. Association for
  Computational Linguistics (2003)

\bibitem{CreutzBaseline}
Creutz, M., Lagus, K.: Unsupervised discovery of morphemes. In: Proceedings of
  the ACL-02 workshop on Morphological and phonological learning. pp. 21--30.
  Association for Computational Linguistics (2002)

\bibitem{CreutzCategoriesMAP}
Creutz, M., Lagus, K.: Inducing the morphological lexicon of a natural language
  from unannotated text. In: Proceedings of the International and
  Interdisciplinary Conference on Adaptive Knowledge Representation and
  Reasoning (AKRR 2005). pp. 106--113 (2005)

\bibitem{Creutz2007}
Creutz, M., Lagus, K.: Unsupervised models for morpheme segmentation and
  morphology learning. ACM Transactions on Speech Language Processing  4,
  1--34 (2007)

\bibitem{Dejean}
D{\'e}jean, H.: Morphemes as necessary concept for structures discovery from
  untagged corpora. In: Proceedings of the Joint Conferences on New Methods in
  Language Processing and Computational Natural Language Learning. pp.
  295--298. Association for Computational Linguistics (1998)

\bibitem{GoldwaterInterpolatingBetween}
Goldwater, S., Johnson, M., Griffiths, T.L.: Interpolating between types and
  tokens by estimating power-law generators. In: Advances in Neural Information
  Processing Systems 18, pp. 459--466. MIT Press (2006)

\bibitem{HaferWeiss}
Hafer, M.A., Weiss, S.F.: Word segmentation by letter successor varieties.
  Information Storage and Retrieval  10(11-12),  371 -- 385 (1974)

\bibitem{Hankamer}
Hankamer, J.: Finite state morphology and left to right phonology. In:
  Proceedings of the West Coast Conference on Formal Linguistics. vol.~5, pp.
  41--52 (1986)

\bibitem{1955Harris}
Harris, Z.S.: From phoneme to morpheme. Language  31(2),  190--222 (1955)

\bibitem{MorphoChallengeWeb}
Kurimo, M., Lagus, K., Virpioja, S., Turunen, V.: Morpho {C}hallenge 2010.
  \url{http://research.ics.tkk.fi/events/morphochallenge2010/} (2011), online;
  accessed 31-January-2017

\bibitem{Mikolov}
Mikolov, T., Chen, K., Corrado, G., Dean, J.: Efficient estimation of word
  representations in vector space. CoRR  abs/1301.3781 (2013),
  \url{http://arxiv.org/abs/1301.3781}

\bibitem{Karthik}
Narasimhan, K., Barzilay, R., Jaakkola, T.S.: An unsupervised method for
  uncovering morphological chains. Transactions of the Association for
  Computational Linguistics  3,  157--167 (2015)

\bibitem{Snyder2008}
Snyder, B., Barzilay, R.: Unsupervised multilingual learning for morphological
  segmentation. In: Proceedings of ACL-08: HLT. pp. 737--745. Association for
  Computational Linguistics (June 2008)

\bibitem{Soricut}
Soricut, R., Och, F.: Unsupervised morphology induction using word embeddings.
  In: Proceedings of the Human Language Technologies: The 2015 Annual
  Conference of the North American Chapter of the ACL. pp. 1627--1637.
  Association for Computational Linguistics (2015)

\bibitem{ustun2016unsupervised}
{\"U}st{\"u}n, A., Can, B.: Unsupervised morphological segmentation using
  neural word embeddings. In: Statistical Language and Speech Processing: 4th
  International Conference, SLSP 2016, Pilsen, Czech Republic, October 11-12,
  2016, Proceedings. pp. 43--53. Springer International Publishing (2016)

\end{thebibliography}
